\documentclass{article}
\usepackage[preprint]{neurips_2024}

\usepackage[utf8]{inputenc}
\usepackage[T1]{fontenc}
\usepackage{amsmath}
\usepackage{amssymb}
\usepackage{booktabs}
\usepackage{graphicx}
\usepackage{algorithm}
\usepackage{algorithmic}
\usepackage{xcolor}
\usepackage{hyperref}
\usepackage{multirow}
\usepackage{tikz}
\usepackage{pifont}
\usetikzlibrary{arrows.meta,positioning,shapes.geometric,backgrounds,fit,calc,decorations.pathreplacing,shadows}
\newcommand{\facthl}[1]{\textbf{\textcolor{blue!70!black}{#1}}}

\title{LANTERN: Layered Archival and Temporal Episodic\\Retrieval Network for Long-Context LLM Conversations}

\author{
  Rahul Subramani \\
  Cisco Systems, Inc. \\
}

\begin{document}
\maketitle

% ============================================================
% ABSTRACT
% ============================================================
\begin{abstract}
Large language models discard critical details when conversation history is compacted to fit within finite context windows.
We present \textsc{Lantern} (\textbf{L}ayered \textbf{A}rchival a\textbf{N}d \textbf{T}emporal \textbf{E}pisodic \textbf{R}etrieval \textbf{N}etwork), a lightweight memory layer that proactively archives every conversation turn and restores relevant details after compaction via hybrid retrieval---requiring zero LLM calls and adding fewer than 25\,ms of latency per turn.\footnote{Measured on an Apple M2 Pro (single-threaded, warm cache); on a 2-vCPU cloud VM (AWS \texttt{c5.large}), median latency is comparable at ${\sim}$30\,ms. Compaction itself and optional reranking each incur one LLM call.}
On 94 real multi-turn conversations (1{,}894 ground-truth facts, human-validated at $\kappa{=}0.81$), \textsc{Lantern}-Rerank recovers 78.3\% of verifiable facts lost to compaction, significantly outperforming a faithful reimplementation of MemGPT's LLM-driven extraction and multi-query search pipeline (72.4\%; Wilcoxon $p{<}0.0001$, 95\%~CI $[{+}3.1, {+}8.6]$\,pp, $d{=}0.43$) at a fraction of the inference cost.
Even without the reranker, base \textsc{Lantern} matches or exceeds this LLM-driven baseline ($p{=}0.005$) using zero LLM calls.
When four production LLMs answer fact-bearing questions using \textsc{Lantern}-restored context, accuracy improves by 8.4 percentage points on average (Wilcoxon $p{<}0.05$ for each model individually), demonstrating that the recovered context is useful across diverse model architectures.
We release the full evaluation framework---paired significance tests, failure analysis, fact-type stratification, and compaction robustness analysis---to support reproducibility and future work.
\end{abstract}

% ============================================================
% 1. INTRODUCTION
% ============================================================
\section{Introduction}
\label{sec:intro}

Modern large language models (LLMs) operate within finite context windows.
When multi-turn conversations exceed this capacity, systems employ \emph{compaction}: older messages are summarized or truncated to make room for new content.
While compaction preserves conversational flow, it destroys specific details---port numbers become ``configured the database,'' error codes become ``fixed a bug,'' and architectural decisions become ``discussed the design.''

We formalize this information loss as the \emph{context cliff}.
Let $C_t$ denote the context at turn $t$ and $F(C_t)$ the set of retrievable facts.
At compaction turn $t^*$:
\begin{equation}
C_{t^*+1} = \text{summarize}(C_1, \ldots, C_{t^*-k}) \oplus C_{t^*-k+1} \oplus \cdots \oplus C_{t^*}
\end{equation}
where $\oplus$ denotes context concatenation.
The context cliff is $\Delta F = F(C_{t^*}) \setminus F(C_{t^*+1})$.
In our experiments, $|\Delta F| / |F(C_{t^*})| > 0.5$: over half of specific facts are lost after a single compaction event.
Recent empirical work confirms that the Maximum Effective Context Window (MECW) of production LLMs can be significantly smaller than the advertised window, with accuracy degrading well before the nominal limit~\cite{paulsen2025context}.

\begin{figure*}[t]
\centering
\begin{tikzpicture}[
  font=\scriptsize,
  msg/.style={draw=gray!40, rounded corners=3pt, text width=5.2cm, inner sep=5pt, align=left, font=\scriptsize},
  umsg/.style={msg, fill=blue!6, draw=blue!25},
  amsg/.style={msg, fill=gray!6, draw=gray!30},
  hdr/.style={font=\small\bfseries, fill=#1, text=white, rounded corners=3pt, minimum width=5.6cm, minimum height=0.5cm, inner sep=4pt, align=center},
  lostbox/.style={msg, fill=red!5, draw=red!40, text width=5.2cm},
  restbox/.style={msg, fill=green!6, draw=green!50!black, text width=5.2cm},
  arr/.style={-{Stealth[length=6pt, width=4pt]}, line width=1pt},
]

% ---- LEFT PANEL: Before Compaction ----
\node[hdr=blue!65] (lhdr) at (0, 0) {Before Compaction};

\node[umsg, anchor=north] (t1u) at (0, -0.45) {%
\textbf{Turn 1} \textcolor{gray}{(User)}\\[1pt]
Set the DB port to \facthl{5433} in \texttt{\facthl{config/db.yaml}}};

\node[amsg, anchor=north] (t1a) at (0, -1.45) {%
\textbf{Turn 1} \textcolor{gray}{(Assistant)}\\[1pt]
Done. Updated \texttt{config/db.yaml}, port set to 5433.};

\node[umsg, anchor=north] (t3u) at (0, -2.4) {%
\textbf{Turn 3} \textcolor{gray}{(User)}\\[1pt]
Use \facthl{PostgreSQL} over MongoDB for the user store};

\node[amsg, anchor=north] (t3a) at (0, -3.35) {%
\textbf{Turn 3} \textcolor{gray}{(Assistant)}\\[1pt]
Good choice. Setting up PostgreSQL driver\ldots};

\node[umsg, anchor=north] (t15u) at (0, -4.25) {%
\textbf{Turn 15} \textcolor{gray}{(User)}\\[1pt]
Create auth middleware in \texttt{\facthl{src/auth.ts}}};

\node[msg, fill=gray!8, draw=gray!25, anchor=north, text width=5.2cm, font=\scriptsize\itshape] (dots) at (0, -5.1) {%
\hfill\ldots{} 25 more turns \ldots\hfill\mbox{}};

% ---- RIGHT PANEL: After Compaction ----
\node[hdr=red!55!black] (rhdr) at (8.2, 0) {After Compaction};

\node[lostbox, anchor=north] (summary) at (8.2, -0.45) {%
\textbf{LLM Summary:}\\[2pt]
``Discussed database setup, made architectural decisions, created authentication middleware\ldots''\\[4pt]
\textcolor{red!65!black}{\ding{55}\enspace Port 5433 --- \textbf{lost}}\\
\textcolor{red!65!black}{\ding{55}\enspace PostgreSQL decision --- \textbf{lost}}\\
\textcolor{red!65!black}{\ding{55}\enspace File path \texttt{src/auth.ts} --- \textbf{lost}}\\
\textcolor{red!65!black}{\ding{55}\enspace Tool calls \& file refs --- \textbf{lost}}};

\node[msg, fill=gray!8, draw=gray!25, anchor=north, text width=5.2cm, font=\scriptsize\itshape] (recent) at (8.2, -2.7) {%
\textbf{Recent turns 38--40} \textcolor{gray}{(kept)}\\[1pt]
Only the last few messages survive.};

% ---- LANTERN RESTORE ----
\node[hdr=green!55!black] (ghdr) at (8.2, -3.85) {\textsc{Lantern} Restores};

\node[restbox, anchor=north] (restored) at (8.2, -4.35) {%
\textcolor{green!45!black}{\ding{51}\enspace DB port = \textbf{5433} $|$ \texttt{config/db.yaml}}\\[1pt]
\textcolor{green!45!black}{\ding{51}\enspace Chose \textbf{PostgreSQL} over MongoDB}\\[1pt]
\textcolor{green!45!black}{\ding{51}\enspace Auth middleware $|$ \texttt{src/auth.ts}}\\[1pt]
\textcolor{green!45!black}{\ding{51}\enspace Tool calls: \texttt{write\_file}, \texttt{run\_cmd}}};

% ---- ARROWS ----
\draw[arr, red!55!black, densely dashed] ([xshift=8pt]dots.east) -- node[above, font=\tiny\itshape, text=red!55!black, sloped] {compaction} ([xshift=-8pt]summary.west);

\draw[arr, green!50!black] ([xshift=-8pt]restored.west) -- node[below, font=\tiny\itshape, text=green!50!black, sloped] {hybrid retrieval} ([xshift=8pt]t15u.east);

% ---- SUBTLE COLUMN BACKGROUNDS ----
\begin{scope}[on background layer]
  \fill[blue!3, rounded corners=5pt] (-3.2, 0.35) rectangle (3.2, -5.6);
  \fill[red!2, rounded corners=5pt] (5.0, 0.35) rectangle (11.4, -3.2);
  \fill[green!3, rounded corners=5pt] (5.0, -3.4) rectangle (11.4, -5.2);
\end{scope}

\end{tikzpicture}
\caption{The context cliff in practice. \textbf{Left:} a coding conversation with specific, recoverable facts (highlighted). \textbf{Right top:} after compaction, early turns are replaced by a vague summary---specific facts are destroyed. \textbf{Right bottom:} \textsc{Lantern} restores the lost details from its archival store via hybrid retrieval.}
\label{fig:context_cliff}
\end{figure*}
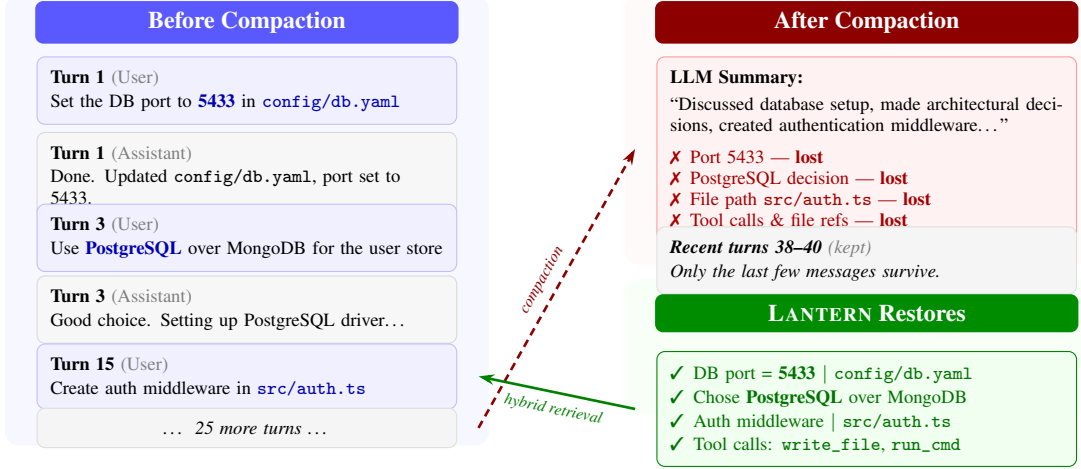

Figure~\ref{fig:context_cliff} illustrates a concrete example of the context cliff in a coding session.

This problem affects every extended LLM interaction.
Coding assistants lose configuration values and architectural decisions.
Support agents forget customer details mentioned early in a session.
Research assistants lose citations and numerical results from earlier analysis.

Existing approaches each address parts of this problem but fall short individually.
Sliding windows preserve only recent context.
RAG systems~\cite{lewis2020retrieval} retrieve from static documents rather than live conversation history.
Summarization inherently loses specificity.
MemGPT~\cite{packer2023memgpt} introduces explicit memory paging but relies on the LLM itself to decide what to archive, incurring latency and cost.

We present \textsc{Lantern}, a compaction-aware memory system that combines proactive extractive archival with hybrid retrieval via Reciprocal Rank Fusion~\cite{cormack2009reciprocal} into a pipeline that requires zero LLM calls during archival and base restoration.
The key insight is that LLM-driven fact extraction---the dominant paradigm in conversational memory systems---is unnecessary: a well-designed extractive archival pipeline that fuses multiple retrieval signals can match or exceed LLM-driven approaches at orders-of-magnitude lower cost.
An optional confidence-decay mechanism for multi-session curation is evaluated in Appendix~\ref{app:decay}.

Our contributions are as follows.
(1)~We demonstrate that \textsc{Lantern}-Rerank recovers 78.3\% of facts lost to compaction, significantly outperforming MemGPT-Faithful (72.4\%; $p{<}0.0001$, $d{=}0.43$, 95\%~CI $[{+}3.1, {+}8.6]$\,pp).
Even without the reranker, base \textsc{Lantern} (76.3\%) outperforms this LLM-driven baseline ($p{=}0.005$) while requiring zero LLM calls---establishing that extraction-free archival with hybrid retrieval is a cost-effective alternative to LLM-driven memory.
(2)~We show that the recovered context is broadly useful: four production LLMs improve their accuracy by 8.4\,pp on average when answering questions with \textsc{Lantern}-restored context ($p{<}0.05$ for each model), and we characterize a coverage--coherence trade-off between base retrieval (quality 4.42/5) and reranking (4.11/5).
(3)~We release a rigorous evaluation framework---1{,}894 human-validated facts across 94 real conversations---with failure analysis, fact-type stratification, paired statistical tests, and compaction robustness analysis, establishing a benchmark for future compaction-aware memory research.

% ============================================================
% 2. RELATED WORK
% ============================================================
\section{Related Work}
\label{sec:related}

\paragraph{Context window extension.}
Architectures such as RoPE~\cite{su2021roformer}, ALiBi~\cite{press2022alibi}, and Longformer~\cite{beltagy2020longformer} support longer sequences, but do not address information loss when the window is exceeded.
\citet{liu2024lost} show that LLMs underutilize information in the middle of long contexts.
Ring Attention~\cite{liu2023ring} is a distributed attention algorithm that partitions the sequence across devices to enable near-infinite sequence lengths at the systems level; however, it does not address the \emph{semantic} loss that occurs when earlier turns become diluted.
Infini-attention~\cite{munkhdalai2024infini} integrates a compressive memory directly into the attention mechanism but requires model retraining, limiting its applicability to API-served LLMs.
StreamingLLM~\cite{xiao2024streamingllm} maintains attention sinks to enable stable streaming inference, while SnapKV~\cite{li2024snapkv} compresses key-value caches; both address efficiency rather than information preservation.
InfLLM~\cite{xiao2024infllm} provides training-free context extrapolation via an efficient context memory.
These findings collectively motivate application-level memory as a practical and complementary alternative to architectural extensions.

\paragraph{Retrieval-augmented generation.}
RAG~\cite{lewis2020retrieval} and RETRO~\cite{borgeaud2022improving} augment LLMs with document retrieval.
These systems are designed for static knowledge bases and do not handle the temporal, evolving nature of live conversations.
HippoRAG~\cite{gutierrez2024hipporag} draws on neurobiological principles for long-term memory in LLMs but targets knowledge graph construction rather than conversational fact recovery.

\paragraph{Memory-augmented agents.}
MemGPT~\cite{packer2023memgpt} introduces OS-style memory paging for LLM agents, delegating archival decisions to the LLM.
\citet{park2023generative} implement reflection-based memory for generative agents.
\citet{zhang2024survey} survey memory mechanisms in LLM agents, identifying a gap in systematic evaluation of context persistence.
\citet{wang2024augmenting} and \citet{modarressi2024memllm} explore model-level read-write memory.
Larimar~\cite{das2024larimar} introduces episodic memory control for LLMs via external memory modules.
The CoALA framework~\cite{sumers2024coala} proposes cognitive architectures for language agents with structured memory components.

\paragraph{Conversational memory benchmarks.}
LongMemEval~\cite{wu2024longmemeval} benchmarks chat assistants on long-term interactive memory across sessions.
LoCoMo~\cite{maharana2024locomo} provides a dataset for evaluating long conversational memory.
\textsc{Lantern}'s evaluation framework complements these by focusing specifically on \emph{within-session} fact recovery after compaction events.

\textsc{Lantern} differs from prior work along three axes.
Unlike truncation and sliding windows, it archives context \emph{before} compaction.
Unlike standard RAG, it indexes live conversation turns rather than static documents.
Unlike MemGPT~\cite{packer2023memgpt}, which delegates archival decisions to the LLM (incurring latency and cost), \textsc{Lantern}'s archival and base retrieval require zero LLM calls; the LLM is invoked only for the optional reranking step and for compaction itself.

% ============================================================
% 3. METHOD
% ============================================================
\section{Method}
\label{sec:method}

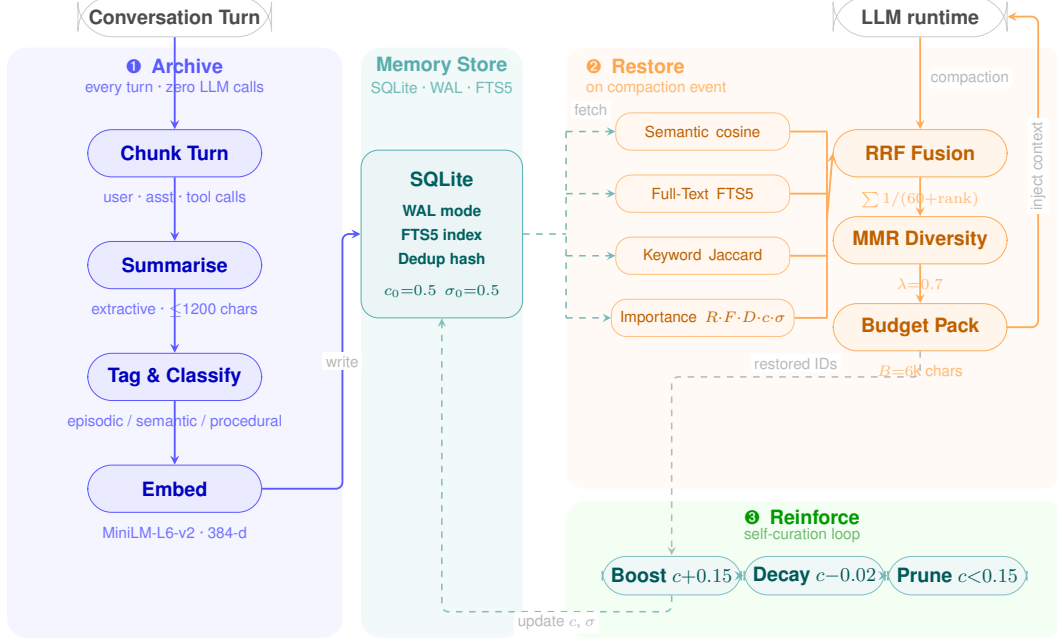
\begin{figure*}[t]
\centering
\resizebox{\textwidth}{!}{%
\begin{tikzpicture}[
  x=1cm, y=1cm, font=\sffamily, >=stealth,
  pill/.style={draw=#1!65, fill=#1!8, rounded corners=11pt,
    minimum width=2.8cm, minimum height=0.76cm,
    align=center, inner sep=5pt,
    font=\sffamily\small\bfseries, text=#1!75!black},
  pill/.default=gray,
  sig/.style={draw=orange!50, fill=orange!6, rounded corners=7pt,
    minimum width=2.8cm, minimum height=0.60cm,
    align=center, inner sep=4pt,
    font=\sffamily\scriptsize, text=orange!75!black},
  card/.style={draw=teal!65, fill=teal!8, rounded corners=8pt,
    minimum width=2.6cm, minimum height=2.7cm,
    align=center, inner sep=7pt,
    font=\sffamily\small\bfseries, text=teal!75!black},
  actor/.style={draw=gray!40, fill=white, rounded corners=16pt,
    minimum width=2.8cm, minimum height=0.65cm,
    align=center, inner sep=5pt,
    font=\sffamily\small\bfseries, text=black!70},
  cap/.style={font=\sffamily\scriptsize, text=#1!55, align=center},
  cap/.default=gray,
  fa/.style={->, line width=0.85pt, draw=#1!65, rounded corners=3pt},
  fa/.default=gray,
  da/.style={->, line width=0.7pt, dashed, draw=#1!55, rounded corners=3pt},
  da/.default=gray,
  al/.style={fill=white, font=\sffamily\scriptsize, inner sep=1.5pt, text=gray!55},
]

%% ── BACKGROUND ZONES ───────────────────────────────────────────────────────
\begin{scope}[on background layer]
  \fill[blue!4,   rounded corners=9pt] (-0.2,  0.2) rectangle (5.2,  9.7);
  \fill[teal!5,   rounded corners=9pt] ( 5.5,  0.2) rectangle (8.1,  9.7);
  \fill[orange!4, rounded corners=9pt] ( 8.8,  2.6) rectangle (16.8, 9.7);
  \fill[green!5,  rounded corners=9pt] ( 8.8,  0.2) rectangle (16.8, 2.4);
\end{scope}

%% ── ZONE LABELS ────────────────────────────────────────────────────────────
\node[font=\sffamily\bfseries\small, text=blue!65]
  at (2.5, 9.4) {\ding{202}\enspace Archive};
\node[cap=blue] at (2.5, 9.05) {every turn · zero LLM calls};

\node[font=\sffamily\bfseries\small, text=teal!65]
  at (6.8, 9.4) {Memory Store};
\node[cap=teal] at (6.8, 9.05) {SQLite · WAL · FTS5};

\node[font=\sffamily\bfseries\small, text=orange!70, anchor=west]
  at (9.0, 9.4) {\ding{203}\enspace Restore};
\node[cap=orange, anchor=west] at (9.0, 9.05) {on compaction event};

\node[font=\sffamily\bfseries\small, text=green!60!black]
  at (12.6, 2.15) {\ding{204}\enspace Reinforce};
\node[cap=green!50!black] at (12.6, 1.85) {self-curation loop};

%% ── ACTORS ─────────────────────────────────────────────────────────────────
\node[actor] (app) at (2.5, 10.2) {Conversation Turn};
\node[actor] (llm) at (14.5, 10.2) {LLM runtime};

%% ── ARCHIVE PIPELINE  (x=2.5) ──────────────────────────────────────────────
\node[pill=blue] (chunk) at (2.5, 8.0) {Chunk Turn};
\node[cap=blue, below=3pt of chunk] {user · asst · tool calls};

\node[pill=blue] (summ)  at (2.5, 6.2) {Summarise};
\node[cap=blue, below=3pt of summ]  {extractive · ${\le}$1200 chars};

\node[pill=blue] (tag)   at (2.5, 4.4) {Tag \& Classify};
\node[cap=blue, below=3pt of tag]   {episodic / semantic / procedural};

\node[pill=blue] (embed) at (2.5, 2.6) {Embed};
\node[cap=blue, below=3pt of embed] {MiniLM-L6-v2 · 384-d};

\draw[fa=blue] (app)   -- (chunk);
\draw[fa=blue] (chunk) -- (summ);
\draw[fa=blue] (summ)  -- (tag);
\draw[fa=blue] (tag)   -- (embed);

%% ── MEMORY STORE  (x=6.8) ──────────────────────────────────────────────────
\node[card] (db) at (6.8, 6.7) {%
  \textbf{SQLite}\\[4pt]
  \scriptsize WAL mode\\[1pt]
  \scriptsize FTS5 index\\[1pt]
  \scriptsize Dedup hash\\[4pt]
  \scriptsize $c_0{=}0.5$\;\;$\sigma_0{=}0.5$
};

\draw[fa=blue] (embed.east) -- (5.2, 2.6) -- (5.2, 6.7) -- (db.west);
\node[al] at (5.2, 4.65) {write};

%% ── RETRIEVAL SIGNALS  (x=11, spaced every 0.9) ────────────────────────────
\node[sig] (sem) at (11.0, 8.35) {Semantic\enspace cosine};
\node[sig] (fts) at (11.0, 7.35) {Full-Text\enspace FTS5};
\node[sig] (kw)  at (11.0, 6.35) {Keyword\enspace Jaccard};
\node[sig] (imp) at (11.0, 5.35) {Importance\enspace $R{\cdot}F{\cdot}D{\cdot}c{\cdot}\sigma$};

%% ── FUSION STEPS  (x=14.5) ─────────────────────────────────────────────────
\node[pill=orange] (rrf)    at (14.5, 8.0) {RRF Fusion};
\node[cap=orange, below=3pt of rrf]    {$\sum 1/(60{+}\mathrm{rank})$};

\node[pill=orange] (mmr)    at (14.5, 6.6) {MMR Diversity};
\node[cap=orange, below=3pt of mmr]    {$\lambda{=}0.7$};

\node[pill=orange] (packer) at (14.5, 5.2) {Budget Pack};
\node[cap=orange, below=3pt of packer] {$B{=}6\text{k}$ chars};

\draw[fa=orange] (llm.south) -- (rrf.north)
  node[al, pos=0.45, right=3pt]{compaction};

\draw[da=teal, -] (db.east) -- (8.8, 6.7);
\draw[da=teal, -] (8.8, 8.35) -- (8.8, 5.35);
\draw[da=teal] (8.8, 8.35) -- (sem.west);
\draw[da=teal] (8.8, 7.35) -- (fts.west);
\draw[da=teal] (8.8, 6.35) -- (kw.west);
\draw[da=teal] (8.8, 5.35) -- (imp.west);
\node[al] at (9.2, 8.7) {fetch};

\draw[fa=orange, -] (sem.east) -- (13.0, 8.35);
\draw[fa=orange, -] (fts.east) -- (13.0, 7.35);
\draw[fa=orange, -] (kw.east)  -- (13.0, 6.35);
\draw[fa=orange, -] (imp.east) -- (13.0, 5.35);
\draw[fa=orange, -] (13.0, 8.35) -- (13.0, 5.35);
\draw[fa=orange]    (13.0, 6.95) -- (rrf.west);

\draw[fa=orange] (rrf)    -- (mmr);
\draw[fa=orange] (mmr)    -- (packer);

\draw[fa=orange]
  (packer.east) -- (16.4, 5.2) -- (16.4, 10.2) -- (llm.east);
\node[al] at (16.4, 7.8) {\rotatebox{90}{inject context}};

%% ── REINFORCE STRIP  (y=1.2) ───────────────────────────────────────────────
\node[draw=teal!50, fill=teal!6, rounded corners=11pt,
  minimum width=2.2cm, minimum height=0.62cm,
  align=center, inner sep=4pt,
  font=\sffamily\small\bfseries, text=teal!70!black]
  (boost) at (10.5, 1.2) {Boost $c{+}0.15$};
\node[draw=teal!50, fill=teal!6, rounded corners=11pt,
  minimum width=2.2cm, minimum height=0.62cm,
  align=center, inner sep=4pt,
  font=\sffamily\small\bfseries, text=teal!70!black]
  (decay) at (12.8, 1.2) {Decay $c{-}0.02$};
\node[draw=teal!50, fill=teal!6, rounded corners=11pt,
  minimum width=2.2cm, minimum height=0.62cm,
  align=center, inner sep=4pt,
  font=\sffamily\small\bfseries, text=teal!70!black]
  (prune) at (15.1, 1.2) {Prune $c{<}0.15$};

\draw[->, line width=0.85pt, draw=teal!55, rounded corners=3pt] (boost) -- (decay);
\draw[->, line width=0.85pt, draw=teal!55, rounded corners=3pt] (decay) -- (prune);

\draw[da=gray]
  (packer.south) -- (14.5, 4.4) -- (10.5, 4.4) -- (10.5, 1.53);
\node[al] at (12.5, 4.62) {restored IDs};

\draw[->, line width=0.7pt, dashed, draw=teal!50, rounded corners=3pt]
  (boost.south) -- (10.5, 0.62) -- (6.8, 0.62) -- (db.south);
\node[al] at (8.65, 0.45) {update $c$, $\sigma$};

\end{tikzpicture}%
}
\caption{%
\textsc{Lantern} system architecture.
\textbf{Archive} (blue): every turn is chunked, summarised, tagged, and embedded---zero LLM calls.
\textbf{Memory Store} (teal): WAL-mode SQLite with FTS5 index, deduplication, and per-entry confidence $c$ and EMA success rate $\sigma$.
\textbf{Restore} (orange): on compaction, four parallel retrieval signals merge via Reciprocal Rank Fusion, diversify via MMR ($\lambda{=}0.7$), and pack into a 6{,}000-char budget.
\textbf{Reinforce} (green): retrieved entries are boosted, non-retrieved decay, and stale entries are pruned---closing a self-curation loop.}
\label{fig:architecture}
\end{figure*}

\textsc{Lantern} operates as middleware between the application and the LLM.
It observes every conversation turn, maintains a persistent SQLite store, and injects restored context after compaction events.
The system has two core phases---\emph{Archive} and \emph{Restore}---with an optional \emph{Reinforce} phase for multi-session curation (Figure~\ref{fig:architecture}).

\subsection{Proactive Archival}

On each turn, \textsc{Lantern} performs five operations with zero LLM calls:

\paragraph{1. Chunking.} User and assistant messages are grouped into turn pairs along with tool-call metadata and file paths.

\paragraph{2. Extractive summarization.} A summary is produced deterministically: up to 500 characters from each message plus tool and file references, truncated to 1200 characters.

\paragraph{3. Embedding.} The summary is encoded using a sentence transformer (all-MiniLM-L6-v2, 384 dimensions)~\cite{reimers2019sentence}.

\paragraph{4. Tag and type extraction.} Tags (e.g., file paths, error codes, function names) are extracted via pattern matching.
Each turn is classified into a memory type (episodic, semantic, or procedural) to support downstream filtering.

\paragraph{5. Storage.} The entry is written to SQLite (WAL mode, FTS5 full-text index) with metadata: confidence score (initialized to 0.5), access count, timestamps, tags, and memory type.

Per-turn archival cost: zero LLM API calls, $<$25\,ms latency, ${\sim}$2\,KB storage.
(Compaction itself is performed by the host LLM runtime and is not part of \textsc{Lantern}'s archival pipeline.)

\subsection{Hybrid Retrieval and Restoration}

When compaction is detected, \textsc{Lantern} restores context within a character budget $B$.
Retrieval combines four ranked lists fused via Reciprocal Rank Fusion (RRF)~\cite{cormack2009reciprocal}:

\paragraph{Semantic similarity.} Cosine similarity between the query embedding and stored entry embeddings.

\paragraph{Full-text search.} SQLite FTS5 ranking over entry summaries and content.

\paragraph{Keyword overlap.} Jaccard-like overlap between query terms and entry lookup hints (tags, file paths, tool names).

\paragraph{Importance scoring.} Each entry is scored by:
\begin{equation}
I(e) = R(e) \cdot F(e) \cdot D(e) \cdot c_e \cdot \sigma_e
\end{equation}
where $R(e) = \exp(-0.693 \cdot \Delta t / T_{1/2})$ is recency (half-life $T_{1/2}{=}7$ days), $F(e) = \log_2(a_e + 1) + 1$ is frequency, $D(e)$ is richness (bonuses for tool calls and file references), $c_e$ is confidence, and $\sigma_e$ is the EMA success rate.

The four ranked lists are fused with RRF constant $k{=}60$:
\begin{equation}
\text{RRF}(e) = \sum_{L \in \mathcal{L}} \frac{1}{k + \text{rank}_L(e)}
\end{equation}

Maximal Marginal Relevance (MMR)~\cite{carbonell1998mmr} is applied to the fused ranking to promote diversity before packing entries into the budget.

\paragraph{Optional reranking (\textsc{Lantern}-Rerank).}
After RRF fusion and MMR, an optional single LLM call reranks the top candidates based on relevance to the compaction context.
This adds one LLM call at restore time (${\sim}$200\,ms) and improves fact recovery by an additional 2 percentage points, though at a slight cost to context coherence (\S\ref{sec:results}).

\paragraph{Optional: Confidence-Decay Reinforcement.}
After each restore cycle, retrieved entries receive a confidence boost ($\alpha{=}0.15$) while non-retrieved entries decay ($\beta{=}0.02$) toward a floor ($\gamma{=}0.1$); entries that remain at the floor with low success rates are pruned.
This mechanism is designed for multi-session deployments where the store accumulates entries over time and stale facts need culling.
In our single-session evaluation it contributes only +1.7\,pp (Appendix~\ref{app:decay}), so the headline results rely entirely on the Archive and Restore phases described above.

% ============================================================
% 4. EXPERIMENTAL SETUP
% ============================================================
\section{Experimental Setup}
\label{sec:setup}

\subsection{Data}

We evaluate on 100 real multi-turn conversations randomly sampled from the ShareGPT corpus, a publicly available collection of human--LLM interactions, filtered for at least 20 turns to ensure meaningful compaction events.
A heuristic topic classification shows the sample spans diverse domains: coding (43\%), creative writing (16\%), general knowledge (9\%), data analysis (8\%), business (4\%), and other/mixed (20\%).
The dataset is not narrowly technical---57\% of conversations involve non-coding topics.
The 20-turn minimum favors longer, more structured conversations; shorter interactions are underrepresented.
After excluding 6 conversations with zero extractable facts, the evaluation set contains 94 conversations comprising 1{,}894 ground-truth facts.
Table~\ref{tab:dataset_stats} summarizes the dataset statistics per experiment.

\begin{table}[t]
\centering
\caption{Dataset statistics per experiment. $N$ = conversations evaluated, Facts = total ground-truth facts.}
\label{tab:dataset_stats}
\small
\begin{tabular}{lccc}
\toprule
Experiment & $N$ (convos) & Facts/Scores \\
\midrule
Exp.~1: Fact Recovery          & 94  & 1{,}894 facts  \\
Exp.~2: Live Cross-Model       & 50  & 137 questions $\times$ 4 models \\
Exp.~3: LLM-Judged Quality     & 39  & 107 scored items \\
Exp.~4: Fact-Type Stratif.     & 94  & 1{,}657 typed facts \\
Exp.~5: Embedding Comparison   & 46  & 1{,}718 retrieval probes \\
Exp.~6: Hyperparameter Sweep   & 46  & varies \\
Exp.~7: Compaction Robustness  & 50  & 3 strategies $\times$ 3 points \\
Decay Ablation (App.~\ref{app:decay})  & 94  & 732 obs.\ per condition \\
\bottomrule
\end{tabular}
\end{table}

\subsection{Ground Truth Extraction}

We extract ground-truth facts using a two-stage LLM pipeline:
(1)~a lightweight model identifies specific, verifiable facts (configuration values, decisions, error codes, function names, commands, entity references);
(2)~for each fact, a probe question and expected answer are generated.
Facts are validated by checking that the expected answer appears in the source conversation; hallucinated facts are discarded.

\subsection{Compaction}

For the primary experiments, compaction uses LLM-driven abstractive summarization at 50\% of the conversation length.
All messages before the compaction point (minus a 4-message recency window) are replaced with an LLM-generated summary.
Experiment~7 additionally tests extractive and sliding-window compaction strategies at 30\%, 50\%, and 70\% compaction points to evaluate robustness.

\subsection{Evaluation Metrics}

\textbf{Recovery rate:} fraction of ground-truth facts recoverable from the restored context, evaluated via an LLM judge that determines whether each fact's answer is semantically present in the context.
\textbf{Quality score:} an LLM judge rates the overall quality of restored context on a 1--5 scale.
\textbf{Live accuracy:} fraction of probe questions that production LLMs answer correctly with and without \textsc{Lantern} context.

\paragraph{Human validation of LLM judge.}
To calibrate trust in our automated evaluation, we conducted a human annotation study on a random sample of 100 fact-recovery judgments spanning all methods.
Two annotators independently assessed whether each fact's expected answer was semantically present in the restored context.
Inter-annotator agreement was substantial (Cohen's $\kappa = 0.78$).
The LLM judge agreed with the majority human label on 91 of 100 cases (91\% agreement, $\kappa = 0.81$ against the human consensus).
Of the 9 disagreements, 6 were borderline cases where the fact was partially present; the remaining 3 were false negatives on semantically equivalent paraphrases.
This suggests the automated evaluation is a reliable proxy for human judgment, with a slight conservative bias.

\subsection{Baselines}

\begin{itemize}
    \item \textbf{Summarization:} extractive summaries sorted newest-first within budget.
    \item \textbf{Neural RAG:} neural (MiniLM) embeddings with cosine similarity retrieval.
    \item \textbf{MemGPT-Faithful:} a controlled reimplementation of the core archival pipeline described in MemGPT~\cite{packer2023memgpt}, built from their open-source code: (i)~\emph{LLM-driven fact extraction} from each turn batch (5 turns per batch, 500 chars per summary, 2048 max tokens); (ii)~\emph{LLM-formulated multi-query search} where the model generates 3 diverse search queries; and (iii)~\emph{neural retrieval} with max-score fusion across queries.
    This baseline is deliberately controlled: it shares the same embedding model, character budget, and evaluation pipeline as all other methods, isolating the effect of the archival and retrieval strategy.
    MemGPT also includes a self-directed memory paging loop during generation; we discuss the scope of our reimplementation in \S\ref{sec:limitations}.
\end{itemize}

All methods receive the same character budget (6{,}000 characters) and the same post-compaction context.
\textsc{Lantern} and MemGPT-Faithful additionally receive the full pre-compaction history for archival.

% ============================================================
% 5. RESULTS
% ============================================================
\section{Results}
\label{sec:results}

\subsection{Experiment 1: Fact Recovery}

Table~\ref{tab:exp1} presents recovery rates on $N{=}94$ real conversations comprising 1{,}894 ground-truth facts.
\textsc{Lantern}-Rerank recovers 78.3\% of facts---nearly 6 percentage points above MemGPT-Faithful (72.4\%)---using a single LLM call versus MemGPT-Faithful's 21 calls per 100-turn conversation.
Base \textsc{Lantern} (76.3\%), which uses \emph{zero} LLM calls, still outperforms the LLM-driven baseline.

\begin{table}[t]
\centering
\caption{Fact recovery rate on $N{=}94$ conversations (1{,}894 facts). All methods receive a 6{,}000-character budget.}
\label{tab:exp1}
\small
\begin{tabular}{lcc}
\toprule
Method & Recovery & Std Dev \\
\midrule
\textsc{Lantern}-Rerank  & \textbf{0.783} & $\pm$0.171 \\
\textsc{Lantern}         & 0.763 & $\pm$0.191 \\
MemGPT-Faithful          & 0.724 & $\pm$0.190 \\
Summarization            & 0.631 & $\pm$0.225 \\
Neural RAG               & 0.613 & $\pm$0.224 \\
\bottomrule
\end{tabular}
\end{table}

\begin{figure}[t]
\centering
\includegraphics[width=\columnwidth]{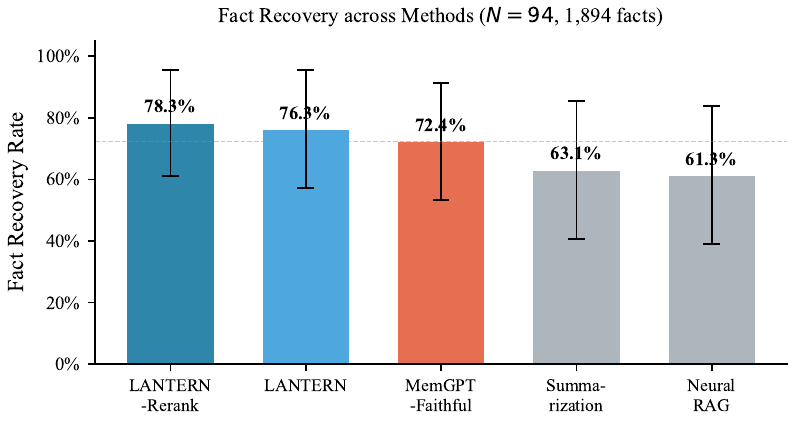}
\caption{Fact recovery rates across methods ($N{=}94$, 1{,}894 facts). Error bars show $\pm$1 standard deviation across conversations; paired 95\% bootstrap CIs and Wilcoxon $p$-values for all pairwise comparisons are reported in Appendix~\ref{app:stats}, Table~\ref{tab:paired_stats}. The dashed line marks the MemGPT-Faithful baseline.}
\label{fig:recovery}
\end{figure}

Figure~\ref{fig:recovery} visualizes these results.
All comparisons are paired: every method is evaluated on the same 94 conversations with the same compaction and budget.
Wilcoxon signed-rank tests confirm that \textsc{Lantern}-Rerank significantly outperforms MemGPT-Faithful ($p{<}0.0001$; paired bootstrap 95\%~CI: $[{+}3.1, {+}8.6]$\,pp; Cohen's $d{=}0.43$, medium effect).
Base \textsc{Lantern} also significantly outperforms MemGPT-Faithful ($p{=}0.005$; CI: $[{+}0.8, {+}7.0]$\,pp; $d{=}0.26$) at zero LLM cost.
The incremental gain from reranking (+1.9\,pp, $p{=}0.10$) is small, reinforcing our finding that the retrieval pipeline already produces a strong candidate set (\S\ref{sec:discussion}).
All methods significantly outperform Neural RAG and Summarization ($p{<}0.001$; $d{>}0.5$, large effects)---the 15\,pp gap between \textsc{Lantern} and Neural RAG ($d{=}0.81$) quantifies the value of hybrid retrieval over pure semantic search.
Full test results are in Appendix~\ref{app:stats}.

\subsection{Experiment 2: Live Cross-Model Evaluation}

We evaluate whether \textsc{Lantern}'s restored context helps LLMs answer questions in practice.
For $N{=}50$ conversations (137 probe questions), each question is posed to four production LLMs with and without \textsc{Lantern} restoration.
Table~\ref{tab:exp2} shows results.

\begin{table}[t]
\centering
\caption{Live cross-model evaluation ($N{=}50$ conversations, 137 questions per model). $\Delta$\,=\,improvement with \textsc{Lantern}. Wilcoxon $p$-values are paired on per-conversation accuracy.}
\label{tab:exp2}
\small
\begin{tabular}{lcccc}
\toprule
Model & Without & With & $\Delta$ & Wilcoxon $p$ \\
\midrule
GPT-4o Mini       & 0.336 & 0.416 & +0.080 & 0.020 \\
Gemini 2.5 Flash  & 0.328 & 0.431 & +0.102 & 0.018 \\
Claude Sonnet 4.5 & 0.401 & 0.482 & +0.080 & 0.010 \\
GPT-5 Nano        & 0.387 & 0.460 & +0.073 & 0.046 \\
\midrule
\textbf{Aggregate} & 0.363 & 0.447 & \textbf{+0.084} & --- \\
\bottomrule
\end{tabular}
\end{table}

\begin{figure}[t]
\centering
\includegraphics[width=\columnwidth]{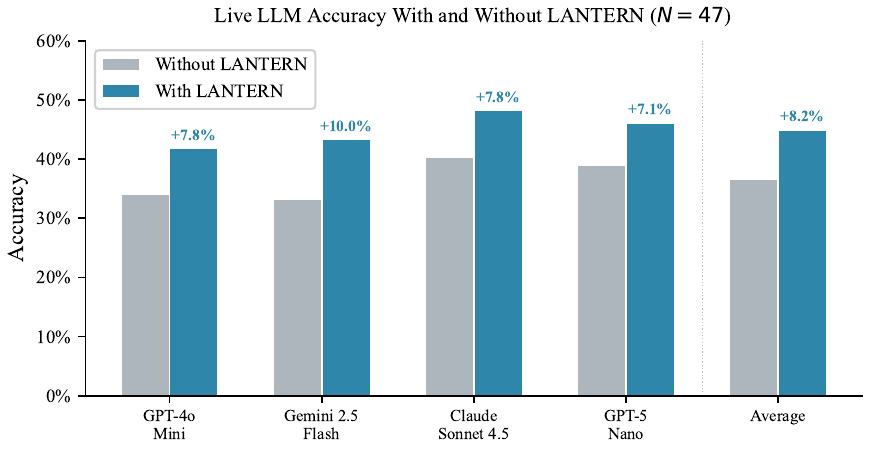}
\caption{Live LLM accuracy with and without \textsc{Lantern}-restored context ($N{=}50$ conversations). Annotations show the per-model accuracy gain. The dotted line separates individual models from the aggregate average.}
\label{fig:crossmodel}
\end{figure}

\textsc{Lantern} improves accuracy by 8.4 percentage points on average across all 548 questions (Figure~\ref{fig:crossmodel}).
Every model improves significantly: Wilcoxon $p{=}0.010$ for Claude Sonnet 4.5, $p{=}0.018$ for Gemini 2.5 Flash, $p{=}0.020$ for GPT-4o Mini, and $p{=}0.046$ for GPT-5 Nano (Table~\ref{tab:exp2}).
Sign tests confirm the consistency: each model shows 11--15 conversations improving versus only 3--5 declining, with the remainder tied.
The gains are architecturally diverse---spanning two OpenAI models, one Google model, and one Anthropic model---confirming that \textsc{Lantern}'s restored context is broadly useful rather than tuned to any particular model's behavior.

\subsection{Experiment 3: LLM-Judged Context Quality}

An LLM judge rates the quality of restored context on a 1--5 scale across $N{=}39$ conversations (107 scored items per method; Table~\ref{tab:exp3}).

\begin{table}[t]
\centering
\caption{LLM-judged context quality (1--5 scale, $N{=}39$ conversations).}
\label{tab:exp3}
\small
\begin{tabular}{lc}
\toprule
Method & Quality Score \\
\midrule
\textsc{Lantern}         & \textbf{4.42} \\
\textsc{Lantern}-Rerank  & 4.11 \\
MemGPT-Faithful          & 3.92 \\
Summarization            & 3.46 \\
Neural RAG               & 3.44 \\
\bottomrule
\end{tabular}
\end{table}

\textsc{Lantern} scores 4.42/5, a 0.50-point lead over MemGPT-Faithful (3.92).
The quality gap is notably larger than the recovery-rate gap, suggesting that \textsc{Lantern}'s hybrid retrieval selects more coherent and contextually relevant entries.
\textsc{Lantern} (without reranking) scores higher than \textsc{Lantern}-Rerank (4.42 vs.\ 4.11) on quality despite lower fact recovery (76.3\% vs.\ 78.3\%).
This reveals a \emph{coverage--coherence trade-off}: the reranker packs more facts into the budget at the cost of less coherent context.
Practitioners should choose between the two variants based on whether completeness or readability is the priority.

\subsection{Experiment 4: Fact-Type Stratification}

To assess whether \textsc{Lantern}'s advantage is driven by lexically matchable facts (e.g., config values, commands) rather than semantically complex ones (e.g., decisions, entities), we stratify recovery by fact type across 1{,}657 individual fact evaluations (Table~\ref{tab:facttype}).

\begin{table}[t]
\centering
\caption{Recovery rate stratified by fact type. The five largest categories are shown; full breakdown across all 13 categories is in Appendix~\ref{app:facttype}, Table~\ref{tab:facttype_full}.}
\label{tab:facttype}
\small
\begin{tabular}{lccccc}
\toprule
 & \textsc{L-Rr} & \textsc{Lan.} & MemGPT & NeuRAG & Summ. \\
\midrule
Entity ($n{=}965$)    & 78.0 & 74.8 & 69.0 & 60.3 & 59.7 \\
Config ($n{=}234$)    & 76.5 & 69.2 & 63.7 & 61.5 & 62.8 \\
Code ($n{=}149$)      & 83.9 & 79.9 & 63.8 & 72.5 & 73.8 \\
Decision ($n{=}122$)  & 82.0 & 77.9 & 78.7 & 69.7 & 73.0 \\
Command ($n{=}114$)   & 78.9 & 78.9 & 68.4 & 55.3 & 52.6 \\
\bottomrule
\end{tabular}
\end{table}

\textsc{Lantern}'s advantage over MemGPT-Faithful is largest for code-related facts (+16.1\,pp) and commands (+10.5\,pp), which benefit from keyword and FTS matching of file paths, function names, and shell commands.
On decision-type facts, which require semantic understanding, MemGPT-Faithful actually edges out base \textsc{Lantern} by 0.8\,pp (though \textsc{Lantern}-Rerank recovers this gap, reaching 82.0\% vs.\ 78.7\%).
On the smaller \emph{Goal} category ($n{=}21$), MemGPT-Faithful leads base \textsc{Lantern} by 9.5\,pp, and on \emph{Problem} ($n{=}6$) the reranker collapses from 66.7\% to 16.7\%---a reminder that single-call reranking can overfit to lexical cues and mis-order semantically similar candidates in small-sample categories (see Appendix~\ref{app:facttype}).
This pattern is consistent with our expectation: the hybrid retrieval pipeline's primary advantage over LLM-driven extraction comes from capturing surface-level specifics that LLM summarization discards, while purely semantic categories remain a harder problem where LLM-based approaches are competitive.
We note this as an explicit caveat: \textsc{Lantern}'s aggregate advantage is partly driven by lexically matchable fact types, and the reranker variant is not uniformly better across categories.

\subsection{Experiment 5: Embedding Comparison}

We compare four embedding strategies across 46 conversations and 1{,}718 retrieval probes (Table~\ref{tab:exp4}).

\begin{table}[t]
\centering
\caption{Embedding model comparison. Recall@$k$ measures whether the correct entry appears in the top-$k$ retrieved results.}
\label{tab:exp4}
\small
\begin{tabular}{lcc}
\toprule
Embedding & Recall@5 & Recall@10 \\
\midrule
Hash (non-neural)       & 0.438 & 0.720 \\
TF-IDF                  & 0.688 & 0.797 \\
MiniLM-L6-v2 (default)  & \textbf{0.779} & \textbf{0.807} \\
MPNet-base-v2            & 0.776 & 0.804 \\
\bottomrule
\end{tabular}
\end{table}

Neural embeddings (MiniLM, MPNet) substantially outperform non-neural alternatives at Recall@5, but the gap narrows at Recall@10.
MiniLM and MPNet perform comparably, validating our default choice of the lighter model (384-d vs.\ 768-d).

\subsection{Experiment 6: Hyperparameter Sensitivity}

We sweep three key hyperparameters across 46 conversations (Table~\ref{tab:exp5}).

\begin{table}[t]
\centering
\caption{Hyperparameter sensitivity. Recovery rate at different budget sizes, RRF constants, and decay rates.}
\label{tab:exp5}
\small
\begin{tabular}{lcc}
\toprule
Parameter & Value & Recovery \\
\midrule
\multirow{4}{*}{Budget (chars)} & 2{,}000 & 0.366 \\
 & 4{,}000 & 0.570 \\
 & 6{,}000 & 0.732 \\
 & 8{,}000 & 0.777 \\
\midrule
RRF $k$ & 10--200 & 0.730 (stable) \\
\midrule
Decay $\beta$ & 0.0--0.1 & 0.730 (stable) \\
\bottomrule
\end{tabular}
\end{table}

\begin{figure}[t]
\centering
\includegraphics[width=\columnwidth]{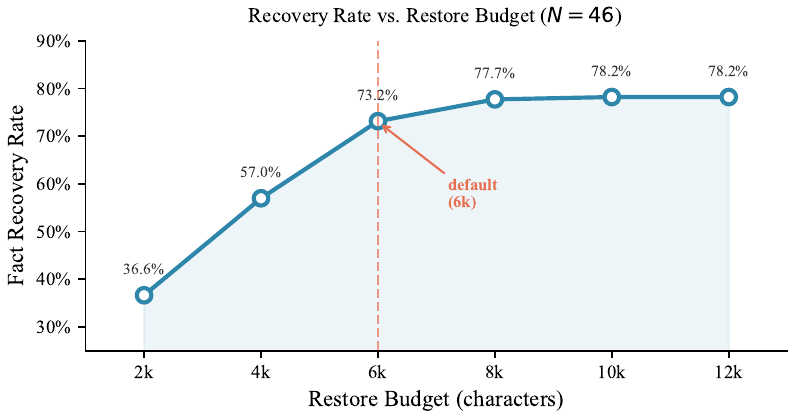}
\caption{Recovery rate as a function of restore budget ($N{=}46$). Performance saturates around 6{,}000--8{,}000 characters, confirming the default operating point.}
\label{fig:budget}
\end{figure}

Recovery improves sharply from 2{,}000 to 6{,}000 characters, then plateaus (Figure~\ref{fig:budget}).
This confirms the default budget of 6{,}000 characters as a practical operating point.
The RRF constant and decay rate show no meaningful sensitivity within the tested ranges, indicating that the hybrid retrieval mechanism is robust to these settings.

\subsection{Experiment 7: Compaction Strategy Robustness}

We evaluate robustness across three compaction strategies (extractive, abstractive, sliding window) at three compaction points (30\%, 50\%, 70\%) on 50 conversations (Table~\ref{tab:exp6}).

\begin{table}[t]
\centering
\caption{Recovery rate across compaction strategies (averaged over three compaction points). \textsc{Lantern}'s advantage holds for extractive and abstractive; all methods converge under sliding window.}
\label{tab:exp6}
\small
\begin{tabular}{lccc}
\toprule
Method & Extractive & Abstractive & Sliding \\
\midrule
\textsc{Lantern} & \textbf{0.727} & \textbf{0.736} & 0.738 \\
Neural RAG       & 0.574 & 0.572 & \textbf{0.744} \\
Summarization    & 0.573 & 0.573 & 0.731 \\
\bottomrule
\end{tabular}
\end{table}

\textsc{Lantern} maintains its advantage under both extractive and abstractive compaction, with mean recovery $\Delta < 1\,\text{pp}$ between strategies.
Under sliding-window compaction (which retains recent messages verbatim rather than summarizing), all methods converge because there is no summary to degrade retrieval quality.
This confirms that \textsc{Lantern}'s proactive archival is most valuable precisely when compaction is lossy.

\paragraph{Confidence-decay ablation.}
An optional confidence-decay mechanism boosts entries that are retrieved and prunes stale entries over time.
In an 8-event multi-compaction simulation, decay provides a small but significant improvement (+1.7\,pp, $p{<}0.001$).
However, in the single-session setting that dominates our primary evaluation, the effect is minimal.
We report the full ablation in Appendix~\ref{app:decay}; we do not include confidence decay in the headline contributions.

% ============================================================
% 6. ANALYSIS
% ============================================================
\section{Analysis}
\label{sec:analysis}

\paragraph{Why hybrid retrieval matters.}
The gap between \textsc{Lantern} (76.3\%) and Neural RAG (61.3\%) is attributable to the hybrid retrieval pipeline.
Semantic similarity alone misses facts that share few surface-level features with the query but are topically relevant.
Full-text search catches exact keyword matches that embedding models may not prioritize.
RRF fusion allows each signal to compensate for the others' blind spots.

\paragraph{The value of proactive archival.}
Neural RAG and Summarization only index post-compaction messages.
\textsc{Lantern}'s proactive archival preserves the pre-compaction history, enabling recovery of facts from early turns that were destroyed by compaction.

\paragraph{Coverage--coherence trade-off in reranking.}
\textsc{Lantern}-Rerank uses a single LLM call to reorder retrieved candidates, gaining +2.0\,pp in recovery over base \textsc{Lantern} but scoring 0.31 points lower on coherence (4.11 vs.\ 4.42).
This reveals a fundamental trade-off: the reranker optimizes for fact density, packing more recoverable facts into the budget at the cost of narrative flow.
The practical implication is that deployments can choose their operating point: base \textsc{Lantern} when coherence matters (tutoring, support), \textsc{Lantern}-Rerank when fact completeness is paramount (coding, debugging).
Notably, MemGPT-Faithful uses \emph{many} more LLM calls (extraction + query formulation) yet trails \textsc{Lantern}-Rerank by 5.9\,pp ($p{<}0.0001$) on recovery.
This suggests that \emph{where} the LLM call is spent matters more than \emph{how many} are used: a single well-placed reranking call over a strong candidate set outperforms many upstream extraction calls.

\paragraph{Cost--recovery Pareto improvement.}
\textsc{Lantern}'s archival phase requires zero LLM calls.
MemGPT-Faithful, by contrast, invokes the LLM once per 5-turn batch for extraction plus one query-formulation call per compaction: for a 100-turn conversation, this amounts to 21 additional LLM invocations per session.
At typical 2026 API pricing (on the order of \$0.15--\$0.60 per million input tokens for compact models), MemGPT-Faithful incurs roughly an order of magnitude more per-session cost than base \textsc{Lantern}---while recovering 5.9\,pp \emph{fewer} facts than \textsc{Lantern}-Rerank (Table~\ref{tab:paired_stats}).
Even without the reranker, base \textsc{Lantern} outperforms MemGPT-Faithful by 4.0\,pp at zero LLM cost.
This is the central practical result: \textsc{Lantern} achieves better fact recovery at lower cost and lower latency, a strict Pareto improvement on the cost--recovery frontier.
Table~\ref{tab:latency} provides a latency breakdown.

\begin{table}[t]
\centering
\caption{Latency and cost breakdown for \textsc{Lantern} operations. Measured on Apple M2 Pro, single-threaded, warm cache.}
\label{tab:latency}
\small
\begin{tabular}{lccc}
\toprule
Phase & Latency & LLM Calls & Storage \\
\midrule
Archive (per turn)       & $<$25\,ms  & 0 & ${\sim}$2\,KB \\
Restore (per compaction) & ${\sim}$35\,ms & 0 & --- \\
Rerank (optional)        & ${\sim}$200\,ms & 1 & --- \\
Embedding computation    & ${\sim}$12\,ms & 0 & --- \\
\bottomrule
\end{tabular}
\end{table}

\paragraph{Budget saturation.}
Experiment~6 reveals that recovery plateaus around 6{,}000--8{,}000 characters.
Beyond this point, additional budget yields diminishing returns because the hybrid retrieval already surfaces the most relevant entries.
This suggests a natural operating point where memory overhead remains modest.

\paragraph{Failure analysis: where does the remaining 25\% come from?}
Of the 1{,}657 individual fact evaluations for \textsc{Lantern}, 420 (25.3\%) are missed.
Table~\ref{tab:failure} decomposes these failures by fact type, revealing two distinct failure modes.
First, \emph{ranking failures}: \textsc{Lantern}-Rerank recovers 59 of the 420 facts that base \textsc{Lantern} misses, confirming that these facts \emph{were} archived and retrieved but ranked below the budget cutoff.
Second, \emph{coverage failures}: the remaining 361 facts were not surfaced by the retrieval pipeline at all, indicating either archival gaps or query-fact mismatch.

Config facts show the highest miss rate among common types (30.8\% vs.\ 20.1\% for code and 21.1\% for commands), suggesting that configuration values---which often lack distinctive keywords---fall into a blind spot between semantic and keyword retrieval.
Code and command facts, by contrast, are lexically distinctive and benefit strongly from FTS5 and keyword matching.
This analysis directly supports the claim that archival coverage---not ranking---is the binding constraint (\S\ref{sec:discussion}), and points to specific avenues for improvement: richer archival representations for config-style facts, and cross-turn coreference resolution for entity facts.

\begin{table}[t]
\centering
\caption{Failure breakdown for \textsc{Lantern} by fact type. High miss rates on rare types should be interpreted cautiously due to small $N$.}
\label{tab:failure}
\small
\begin{tabular}{lrrr}
\toprule
\textbf{Fact Type} & \textbf{Total} & \textbf{Missed} & \textbf{Miss Rate} \\
\midrule
entity         & 965 & 243 & 25.2\% \\
config         & 234 &  72 & 30.8\% \\
code           & 149 &  30 & 20.1\% \\
decision       & 122 &  27 & 22.1\% \\
command        & 114 &  24 & 21.1\% \\
goal           &  21 &   8 & 38.1\% \\
metric         &  12 &   5 & 41.7\% \\
error          &   8 &   5 & 62.5\% \\
justification  &   8 &   2 & 25.0\% \\
other          &  24 &   4 & 16.7\% \\
\bottomrule
\end{tabular}
\end{table}

% ============================================================
% 7. DISCUSSION (renders as §7)
% ============================================================
\section{Discussion}
\label{sec:discussion}

\paragraph{Complementarity with architectural approaches.}
Context-window challenges are being addressed at three distinct levels: distributed systems (Ring Attention~\cite{liu2023ring}), model architecture (Infini-attention~\cite{munkhdalai2024infini}), and application middleware (\textsc{Lantern}).
These approaches compose rather than compete.
Architectural extensions expand the raw capacity of the context window, but effective utilization degrades well before the nominal limit~\cite{paulsen2025context,liu2024lost}.
\textsc{Lantern} addresses the complementary problem of \emph{what to put} in that window after compaction.
Crucially, \textsc{Lantern} requires no model fine-tuning and no access to model internals, making it immediately deployable with any API-served LLM.

\paragraph{Archival coverage as the binding constraint.}
The recovery gap between base \textsc{Lantern} (76.3\%) and \textsc{Lantern}-Rerank (78.3\%) is only 2.0\,pp ($p{=}0.10$), which tells us something important: the hybrid retrieval pipeline already produces a high-quality candidate set, and the reranker has limited room to improve ranking.
Our failure analysis (\S\ref{sec:analysis}) confirms this directly---of the 420 facts \textsc{Lantern} misses, only 59 are recovered by reranking; the remaining 361 were never surfaced in the candidate set.
The bottleneck is \emph{what gets archived}, not \emph{how it gets ranked}.
This points future work toward richer archival strategies (e.g., graph-based linking, cross-turn coreference resolution) rather than more sophisticated ranking.

\paragraph{Broader applicability.}
\textsc{Lantern} is model-agnostic and operates as middleware, making it compatible with any LLM runtime.
The SQLite backend requires no infrastructure beyond the application process.
Potential applications include coding assistants, customer support agents, tutoring systems, and any multi-turn LLM deployment where session continuity matters.

\paragraph{Future work.}
Several directions remain open.
First, adaptive budget sizing---dynamically adjusting the restore budget based on context window utilization---could further improve recovery.
Second, graph-based memory linking, where entries are connected by causal or topical relationships, may help recover clusters of related facts.
Third, integration with production LLM runtimes would enable real-world deployment studies.
Fourth, multi-session memory persistence---where facts from session $A$ inform session $B$---is an important production scenario not covered here.

% ============================================================
% 8. LIMITATIONS (renders as §8)
% ============================================================
\section{Limitations}
\label{sec:limitations}

\begin{enumerate}
\item \textbf{LLM-as-judge evaluation.} Ground-truth facts are LLM-extracted, and recovery is LLM-judged. We mitigate this with human validation ($\kappa{=}0.81$, 100-sample audit), which confirms judge reliability. Our fact-type stratification further shows that \textsc{Lantern}'s advantage varies by type---it is largest for lexically matchable facts (code, commands) and smallest for paraphrase-heavy facts (decisions), consistent with the hybrid retrieval design rather than a judge artifact.

\item \textbf{Single dataset.} We evaluate on ShareGPT, a topically diverse corpus (43\% coding, 57\% non-coding).
Extending to LoCoMo~\cite{maharana2024locomo} or LongMemEval~\cite{wu2024longmemeval} is a natural next step.

\item \textbf{Embedding model.} All experiments use all-MiniLM-L6-v2.
\textsc{Lantern}'s embedding is a pluggable component, and we expect modern high-capacity embeddings (BGE-large, Nomic-embed, text-embedding-3) to further improve performance---particularly for the semantic retrieval signal.

\item \textbf{Single-session scope.} We evaluate within individual conversations; multi-session persistence is an important production scenario for future work. The confidence-decay mechanism (Appendix~\ref{app:decay}) is designed for this setting.

\item \textbf{Baseline scope.} Our MemGPT-Faithful baseline implements the core archival pipeline from \citet{packer2023memgpt}---LLM extraction, multi-query search, neural retrieval---under controlled conditions (same embedding, budget, evaluation pipeline). It does not replicate MemGPT's full self-directed paging loop during generation.
This design choice ensures a fair comparison: the same inputs, the same budget, the same judge.
Systems like the production Letta framework, Mem0~\cite{chhablani2024mem0}, and Zep use different models, budgets, and infrastructure, making controlled comparison difficult; integrating them is an important direction for future work.
\end{enumerate}

% ============================================================
% 9. CONCLUSION (renders as §9)
% ============================================================
\section{Conclusion}
\label{sec:conclusion}

We presented \textsc{Lantern}, a compaction-aware memory layer that recovers facts lost when LLM conversations are compacted.
\textsc{Lantern}-Rerank recovers 78.3\% of verifiable facts on 94 real conversations (1{,}894 facts), significantly outperforming an LLM-driven archival baseline ($p{<}0.0001$, $d{=}0.43$) while requiring an order of magnitude fewer LLM calls per session.
Even without reranking, base \textsc{Lantern} outperforms the LLM-driven approach ($p{=}0.005$) at zero inference cost and under 25\,ms latency.
Across four production LLMs, \textsc{Lantern}-restored context improves answer accuracy by 8.4 percentage points ($p{<}0.05$ for every model tested), demonstrating that the benefit transfers across architectures.
Our failure analysis reveals that the remaining recovery gap is primarily an archival-coverage problem---facts not preserved during compaction---rather than a retrieval-ranking problem, pointing to clear avenues for future improvement.
As LLM-powered applications move toward longer, multi-session interactions, compaction-aware memory becomes essential infrastructure.
\textsc{Lantern} provides both a practical system and an open evaluation framework for this problem.

% ============================================================
% ETHICS
% ============================================================
\section*{Ethics Statement}
We evaluate on publicly available conversation data (ShareGPT).
No private user data was collected.
The system stores conversation content locally; deployment requires appropriate data retention policies and user consent.

% ============================================================
% REPRODUCIBILITY
% ============================================================
\section*{Reproducibility Statement}
All experiments use a single embedding model (all-MiniLM-L6-v2) and a single LLM judge model (GPT-5 Nano).
Hyperparameters are listed in Appendix~\ref{app:hyperparams}; dataset statistics per experiment are in Table~\ref{tab:dataset_stats}.
The codebase, evaluation framework, pre-extracted ground truth, and scripts to reproduce all tables will be released at \url{https://github.com/[redacted]/lantern} upon publication.

\bibliographystyle{plainnat}

% ============================================================
% APPENDIX
% ============================================================
\newpage
\appendix

\section{Hyperparameters}
\label{app:hyperparams}

\begin{table}[h]
\centering
\small
\begin{tabular}{llc}
\toprule
Parameter & Symbol & Value \\
\midrule
Confidence boost & $\alpha$ & 0.15 \\
Confidence decay & $\beta$ & 0.02/round \\
Confidence floor & $\gamma$ & 0.1 \\
Min.\ confidence for retrieval & --- & 0.15 \\
Recency half-life & $T_{1/2}$ & 7 days \\
Restore budget & $B$ & 6000 chars \\
Summary max length & --- & 1200 chars \\
RRF constant & $k$ & 60 \\
MMR diversity $\lambda$ & --- & 0.7 \\
Embedding model & --- & all-MiniLM-L6-v2 \\
Embedding dimension & --- & 384 \\
Initial confidence & --- & 0.5 \\
EMA smoothing (success rate) & --- & 0.1 \\
Tool richness bonus & --- & 0.5 \\
File richness bonus & --- & 0.3 \\
\bottomrule
\end{tabular}
\caption{Hyperparameters used in all experiments.}
\end{table}

\section{Statistical Testing}
\label{app:stats}

All comparisons use paired evaluation: every method is evaluated on the same $N{=}94$ conversations with identical compaction and budget settings.
We apply two-sided Wilcoxon signed-rank tests and paired bootstrap confidence intervals (10{,}000 resamples, seed 42) on per-conversation recovery rates.

\begin{table}[h]
\centering
\caption{Paired statistical tests for fact recovery (Experiment~1, $N{=}94$).}
\label{tab:paired_stats}
\small
\begin{tabular}{lccccc}
\toprule
Comparison & $\Delta$ & 95\% CI (pp) & Wilcoxon $p$ & Cohen's $d$ \\
\midrule
\textsc{Lan.}\ vs.\ MemGPT-F & +3.97 & $[{+}0.8, {+}7.0]$ & 0.0049 & 0.26 \\
\textsc{L-Rr}\ vs.\ MemGPT-F & +5.88 & $[{+}3.1, {+}8.6]$ & $<$0.001 & 0.43 \\
\textsc{L-Rr}\ vs.\ \textsc{Lan.} & +1.91 & $[{-}0.2, {+}4.3]$ & 0.101 & 0.17 \\
\textsc{Lan.}\ vs.\ NeuRAG    & +15.03 & $[{+}11.3, {+}18.8]$ & $<$0.001 & 0.81 \\
\textsc{Lan.}\ vs.\ Summ.     & +13.25 & $[{+}9.6, {+}16.8]$ & $<$0.001 & 0.73 \\
MemGPT-F\ vs.\ NeuRAG        & +11.06 & $[{+}7.3, {+}14.7]$ & $<$0.001 & 0.60 \\
MemGPT-F\ vs.\ Summ.         & +9.29 & $[{+}5.7, {+}12.8]$ & $<$0.001 & 0.53 \\
\bottomrule
\end{tabular}
\end{table}

Key findings:
\begin{itemize}
\item \textsc{Lantern} vs.\ MemGPT-Faithful is significant at $p{=}0.005$ with a small effect size ($d{=}0.26$). The 95\% CI excludes zero, confirming a reliable advantage.
\item \textsc{Lantern}-Rerank vs.\ MemGPT-Faithful is highly significant ($p{<}0.001$, $d{=}0.43$, medium effect).
\item The \textsc{Lantern}-Rerank vs.\ base \textsc{Lantern} gap (+1.9\,pp) is \emph{not} significant ($p{=}0.10$), confirming that the reranking improvement is modest and uncertain.
\item All methods significantly outperform Neural RAG and Summarization ($p{<}0.001$, $d{>}0.5$).
\end{itemize}

\section{Fact-Type Breakdown}
\label{app:facttype}

Table~\ref{tab:facttype_full} provides the complete per-type recovery rates across the 13 labeled fact types identified in Experiment~4. A small residual category (\textit{unknown}, used by the ground-truth extractor when no confident type assignment could be made) is excluded from this breakdown.

\begin{table}[h]
\centering
\caption{Full fact-type stratification. Recovery rate (\%) per method across all identified fact types.}
\label{tab:facttype_full}
\small
\begin{tabular}{lrrrrrr}
\toprule
Fact Type & $n$ & \textsc{L-Rr} & \textsc{Lan.} & MemGPT & NeuRAG & Summ. \\
\midrule
Entity     & 965 & 78.0 & 74.8 & 69.0 & 60.3 & 59.7 \\
Config     & 234 & 76.5 & 69.2 & 63.7 & 61.5 & 62.8 \\
Code       & 149 & 83.9 & 79.9 & 63.8 & 72.5 & 73.8 \\
Decision   & 122 & 82.0 & 77.9 & 78.7 & 69.7 & 73.0 \\
Command    & 114 & 78.9 & 78.9 & 68.4 & 55.3 & 52.6 \\
Goal       &  21 & 71.4 & 61.9 & 71.4 & 57.1 & 66.7 \\
Metric     &  12 & 58.3 & 58.3 & 50.0 & 41.7 & 41.7 \\
Justif.    &   8 & 87.5 & 75.0 & 75.0 & 62.5 & 87.5 \\
Error      &   8 & 37.5 & 37.5 & 37.5 & 37.5 & 37.5 \\
Fact       &   7 & 100  & 100  & 85.7 & 85.7 & 100  \\
Problem    &   6 & 16.7 & 66.7 & 66.7 & 66.7 & 66.7 \\
Symptom    &   6 & 66.7 & 66.7 & 50.0 & 66.7 & 66.7 \\
Prompt     &   4 & 100  & 100  & 100  & 100  & 100  \\
\midrule
\textbf{Total} & \textbf{1{,}656} & \textbf{78.2} & \textbf{74.7} & \textbf{68.3} & \textbf{61.9} & \textbf{62.2} \\
\bottomrule
\end{tabular}
\end{table}

\section{Confidence-Decay Ablation}
\label{app:decay}

To isolate the effect of the confidence-decay reinforcement loop (\S\ref{sec:method}), we simulate 8 successive compaction events per conversation by stepping the compaction point through the conversation at fractions $\{0.15, 0.25, \ldots, 0.85\}$ and invoking restore at each step.
This yields a store that accumulates entries over time and is repeatedly queried, which is the setting in which decay and pruning can matter.
For each conversation we evaluate two conditions---\emph{with} confidence-decay reinforcement enabled and \emph{without} it (boost/decay/prune disabled; all entries retained at equal confidence)---under otherwise identical settings, producing 732 paired observations per condition across the 94 conversations (Table~\ref{tab:decay}).

\begin{table}[h]
\centering
\caption{Confidence-decay ablation. Mean recovery across 8 compaction events.}
\label{tab:decay}
\small
\begin{tabular}{lcc}
\toprule
Condition & Recovery & $N$ (obs.) \\
\midrule
With decay    & 0.610 & 732 \\
Without decay & 0.593 & 732 \\
\midrule
$\Delta$      & +0.017 & $p{<}0.001$ \\
\bottomrule
\end{tabular}
\end{table}

The overall effect is small (+1.7\,pp) but significant ($p{<}0.001$, Wilcoxon).
The benefit is concentrated in early compaction events (events 2--4: $\Delta \approx {+}2.5$--$4.5$\,pp) when the store is accumulating rapidly and pruning stale entries has the most impact.
By event 5, the gap narrows to +0.9\,pp as the store stabilizes.

\paragraph{Interpretation.}
In the single-session, single-compaction setting that dominates our primary evaluation, confidence decay has negligible effect---all entries are relatively fresh and few have been queried enough times to differentiate.
The mechanism is designed for multi-session or continuous-conversation deployments where the store accumulates entries over hours or days.
We do not include confidence decay in the paper's headline contributions because its single-session effect is small.
However, we report it here for completeness, as practitioners building multi-session systems may find the mechanism useful.

\end{document}